\documentclass[10pt,twocolumn,letterpaper]{article}

\usepackage{iccv}
\usepackage{times}
\usepackage{epsfig}
\usepackage{graphicx}
\usepackage{amsmath}
\usepackage{amssymb}
\usepackage{booktabs}
\usepackage{multirow} 
\usepackage{makecell}
\usepackage{rotating}
\usepackage{amsmath}
\DeclareMathOperator*{\argmax}{arg\,max}
\usepackage{wrapfig}
\usepackage{subcaption}
\usepackage[table,xcdraw]{xcolor}

\usepackage{arydshln}
\usepackage{authblk}

\let\svthefootnote\thefootnote
\newcommand\blankfootnote[1]{%
  \let\thefootnote\relax\footnotetext{#1}%
  \let\thefootnote\svthefootnote%
}

\usepackage[pagebackref=true,breaklinks=true,colorlinks,bookmarks=false]{hyperref}

\definecolor{Gray}{gray}{0.95}
\newcolumntype{g}{>{\columncolor{Gray}}c}

\iccvfinalcopy 

\ificcvfinal\fi

\begin{document}

\title{LiDARFormer: A Unified Transformer-based Multi-task Network for LiDAR Perception}

\author[1,2]{\textbf{Zixiang~Zhou}\thanks{Work done during an internship at TuSimple.}\thanks{Contributed equally.}}
\newcommand\CoAuthorMark{\footnotemark[\arabic{footnote}]} 
\author[1]{\textbf{Dongqiangzi~Ye}\protect\CoAuthorMark}
\author[1]{\textbf{Weijia~Chen}}
\author[1]{\textbf{Yufei~Xie}}
\author[1]{\par\textbf{Yu~Wang}}
\author[1]{\textbf{Panqu~Wang}}
\author[2]{\textbf{Hassan~Foroosh}}
\affil[1]{TuSimple}
\affil[2]{University of Central Florida}

\maketitle
\ificcvfinal\thispagestyle{empty}\fi

\begin{abstract}
    There is a recent trend in the LiDAR perception field towards unifying multiple tasks in a single strong network with improved performance, as opposed to using separate networks for each task. In this paper, we introduce a new LiDAR multi-task learning paradigm based on the transformer. The proposed \textbf{LiDARFormer} utilizes cross-space global contextual feature information and exploits cross-task synergy to boost the performance of LiDAR perception tasks across multiple large-scale datasets and benchmarks. Our novel transformer-based framework includes a cross-space transformer module that learns attentive features between the 2D dense Bird's Eye View (BEV) and 3D sparse voxel feature maps. Additionally, we propose a transformer decoder for the segmentation task to dynamically adjust the learned features by leveraging the categorical feature representations. Furthermore, we combine the segmentation and detection features in a shared transformer decoder with cross-task attention layers to enhance and integrate the object-level and class-level features. LiDARFormer is evaluated on the large-scale nuScenes and the Waymo Open datasets for both 3D detection and semantic segmentation tasks, and it outperforms all previously published methods on both tasks. Notably, LiDARFormer achieves the state-of-the-art performance of $76.4\%$ L2 mAPH and $74.3\%$ NDS on the challenging Waymo and nuScenes detection benchmarks for a single model LiDAR-only method.
\end{abstract}

\section{Introduction}

\begin{figure}[th]
  \begin{center}
    \includegraphics[width=0.95\linewidth]{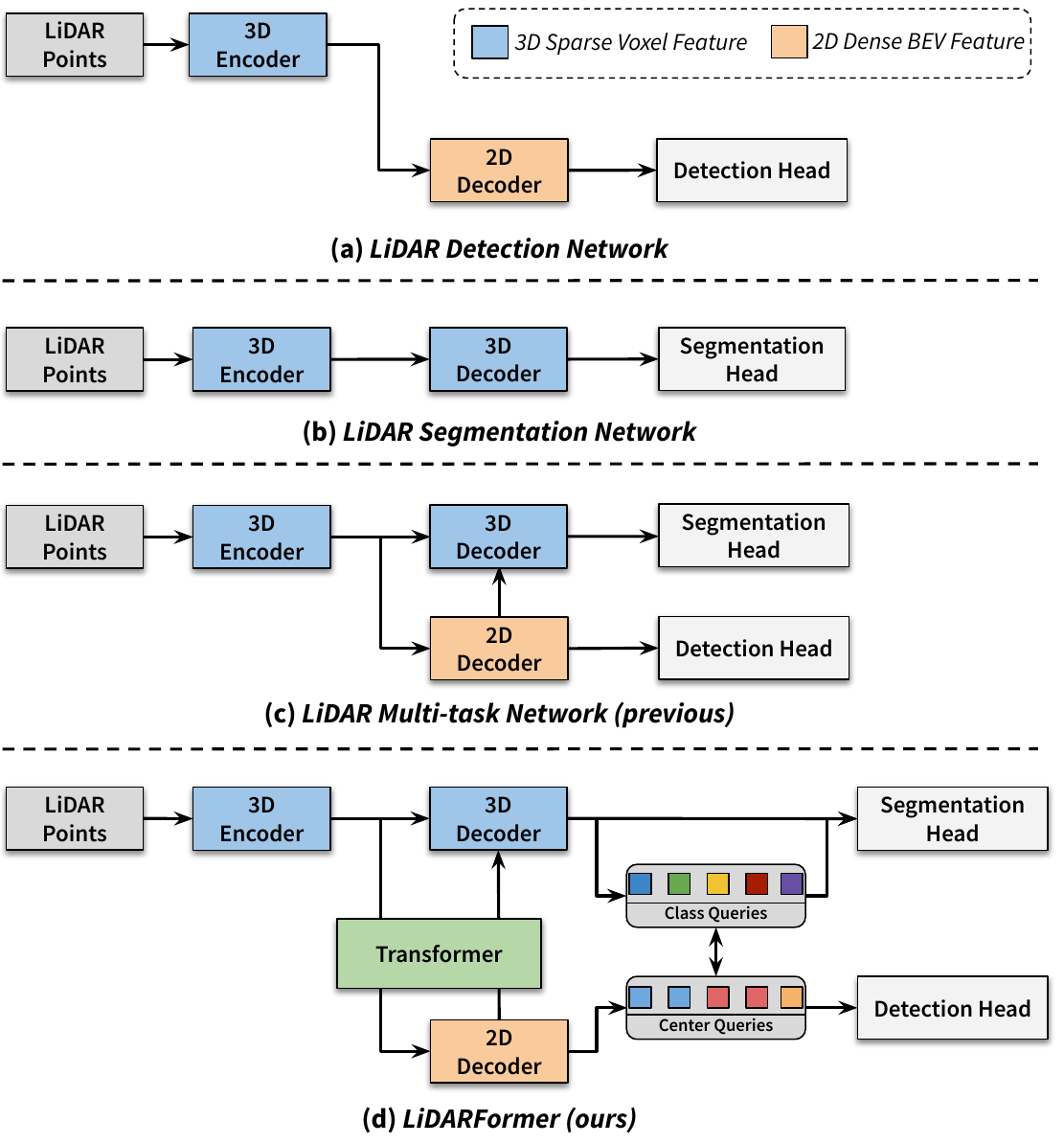}
  \end{center}
  \caption{\textbf{LiDAR Perception Network Designs.} LiDAR detection (a) and segmentation (b) networks typically extract feature representations on distinct feature maps. While a recent multi-task network~\cite{ye2022lidarmultinet} (c) integrates these tasks into a single network, it often overlooks differences among feature maps and the higher-level connections between tasks. Our network (d) utilizes transformer attention to establish more effectively the transformations between 3D sparse and 2D dense features. Moreover, the cross-task information is further shared through class-level and object-level feature embeddings in the multi-task transformer decoder.}
  \label{fig:teaser}
  \vspace{-10pt}
\end{figure}

LiDAR point cloud detection and semantic segmentation tasks aim to predict the object-level 3D bounding boxes and point-level semantic labels, which are among the most fundamental tasks in autonomous vehicle perception. With the recent release of the large-scale LiDAR point cloud datasets~\cite{caesar2020nuscenes,sun2020scalability}, there has been a surge of interest in integrating these tasks into a single framework. Current methods~\cite{lidarmtl2021,ye2022lidarmultinet} rely on voxel-based networks with sparse convolution~\cite{yan2018second,choy20194d} for leading performance. However, different tasks are only connected through sharing the same low-level features without considering the high-level contextual information that is highly related among those tasks. On the other hand, more recent works~\cite{shi2020pv,tang2020searching,xu2021rpvnet} try to fuse features from multiple views that contain both voxel-level and point-level information. These approaches focus more on exploiting local point geometric relations to recover fine-grained details. The problem of efficiently extracting and sharing global contextual information in LiDAR perception tasks is still by and large underexplored.

Meanwhile, transformer-based network structures~\cite{carion2020end,wang2021max,cheng2021maskformer,xie2021segformer,zhang2022dino} start to exhibit an outstanding performance on 2D image detection and segmentation tasks. Apart from directly replacing the conventional CNN with the transformer encoder~\cite{dosovitskiy2020vit,Liu2021SwinTH}, various methods~\cite{carion2020end,zhu2021deformable,cheng2021maskformer,li2022mask} explore using the transformer decoder to extract objects or class-level feature representations, which are served as strong contextual information for feature learning. This transformer decoder design is then adopted in recent LiDAR perception methods~\cite{Zhou_centerformer,bai2022transfusion,marcuzzi2023mask}. However, the transformer decoders used for LiDAR detection and segmentation tasks are performed independently on different feature maps and are not yet unified.

Is it possible to develop a unified transformer-based multi-task LiDAR perception network with the ability to learn global context information? To accomplish this goal, we introduce three novel components in a voxel-based framework. The first component is a cross-space transformer module that enhances the feature mapping between the 3D sparse voxel space and the 2D dense BEV space. These two spaces are frequently used to obtain feature representations for segmentation and detection tasks, respectively. Second, we propose a transformer-based refinement module as the segmentation decoder. The module uses a transformer to extract class feature embeddings and refine voxel features through bidirectional cross-attention. Lastly, we propose a multi-task learning structure that combines segmentation and detection transformer decoders into a unified transformer decoder. By doing so, the network can transfer high-level features through cross-task attention, as depicted in Figure~\ref{fig:teaser}. These three innovative components result in a powerful network, named \textbf{LiDARFormer}, for the next generation of LiDAR perception.

We evaluate our method on two challenging large-scale LiDAR datasets: the nuScenes dataset~\cite{caesar2020nuscenes} and the Waymo Open Dataset~\cite{sun2020scalability}. Our method sets new state-of-the-art standards both in detection and semantic segmentation, by achieving $74.3\%$ NDS on the nuScenes 3D detection and $81.5\%$ mIoU on the nuScenes semantic segmentation. LiDARFormer also achieves $76.4\%$ mAPH in the Waymo Open Dataset detection set, surpassing thus all previous methods. 

Our main contributions are summarized as follows:

\begin{itemize}
    \item We propose a cross-space transformer module to improve feature learning when transferring features between sparse voxel features and dense BEV features in the multi-task network.
    \item We present the first LiDAR cross-task transformer decoder that bridges the information learned across object-level and class-level feature embedding. 
    \item We introduce a transformer-based coarse-to-fine network that utilizes a transformer decoder to extract class-level global contextual information for the LiDAR semantic segmentation task. 
    \item Our network achieves state-of-the-art 3D detection and semantic segmentation performances on two popular large-scale LiDAR benchmarks.
\end{itemize}
\section{Related Work}
\textbf{Voxel-based LiDAR Point Cloud Perception}
Unlike most point cloud networks~\cite{qi2017pointnet,qi2017pointnet++,li2018pointcnn,wu2019pointconv,hu2019randla,xu2020grid,thomas2019KPConv,zhao2021point} that directly learn point-level features in outdoor or indoor point cloud data, LiDAR point cloud perception usually requires transforming the large-scale sparse point cloud into either a 3D voxel map~\cite{zhou2018voxelnet,zhu2021cylindrical}, 2D BEV~\cite{yang2018pixor,lang2019pointpillars,Zhang_2020_CVPR}, or range-view map~\cite{sun2021rsn,fan2021rangedet,wu2018squeezeseg,wu2019squeezesegv2,milioto2019rangenet++,cortinhal2020salsanext}. Thanks to the development of the 3D sparse convolution layer~\cite{yan2018second,choy20194d} in point cloud processing, voxel-based methods are becoming dominant in terms of both high performance and efficient runtime. CenterPoint~\cite{yin2021center} and AFDet~\cite{ge2020afdet} adopted the anchor-free design that detects objects through heatmap classification. Cylinder3D~\cite{zhu2021cylindrical} utilized the cylindrical voxel partition to extract the voxel-level features. LargeKernel3D~\cite{chen2022scaling} showed that the long-range information from a bigger receptive field can significantly improve the performance. LidarMultiNet~\cite{ye2022lidarmultinet} presented a multi-task learning network that unifies different LiDAR perception tasks.

\begin{figure*}[t]
  \begin{center}
    \includegraphics[width=1.0\textwidth]{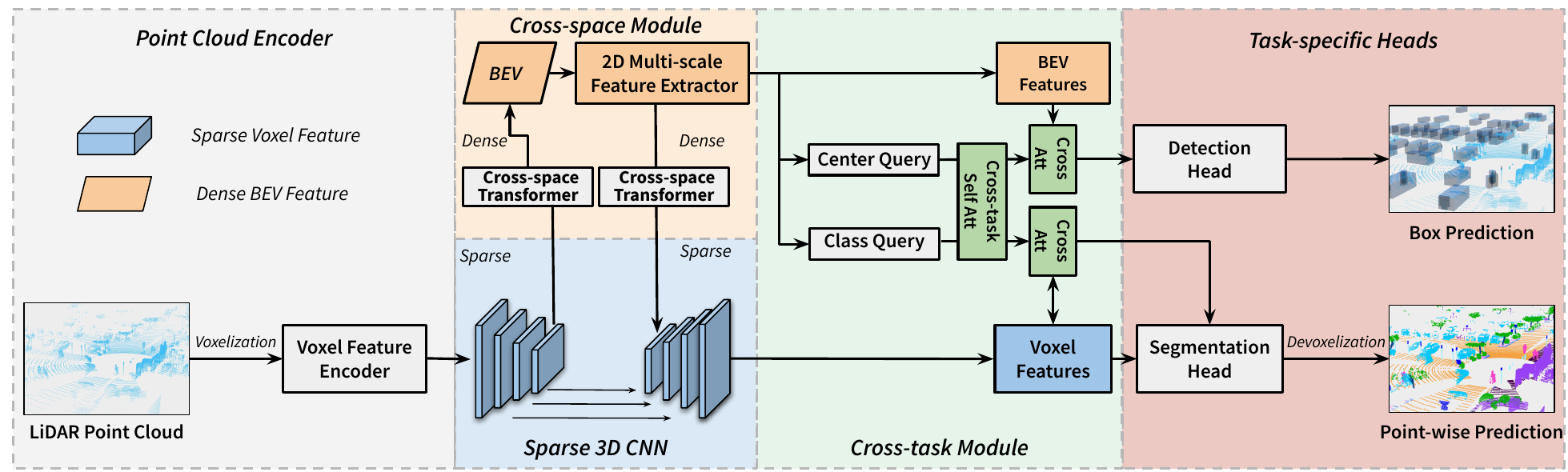}
  \end{center}
  \caption{\textbf{The architecture of LiDARFormer.} Our network first transforms the point cloud into a sparse voxel map. Next, sparse 3D CNN is used to extract voxel feature representation. Between the encoder and the decoder, we use a Cross-space Transformer (\textbf{XSF}) module to learn long-range information in the BEV map. Additionally, we use a cross-task transformer decoder (\textbf{XTF}) to extract class-level and object-level feature representations, which are fed into task-specific heads to generate the detection and segmentation predictions.}
  \label{fig:architecture}
\end{figure*}

Voxel-based methods have to make a trade-off between accuracy and complexity due to the information loss introduced during the projection or voxelization. To alleviate the quantization error, some recent methods~\cite{tang2020searching,shi2020pv,ye2021drinet,xu2021rpvnet} propose to fuse features from multi-view feature maps, combining point-level information with 2D BEV/range-view and 3D voxel features. PVRCNN~\cite{shi2020pv} and SPVNAS~\cite{tang2020searching} used two concurrent point-level and voxel-level feature encoding branches, where these two features were connected at each network block. RPVNet~\cite{xu2021rpvnet} further combined all point, voxel, and range image features in an encoder-decoder segmentation network through a gated fusion module. In contrast to these methods that focus on fine-grained features for details, our method aims to enhance global feature learning in the voxel-based network.

\textbf{Segmentation Refinement} 
In the image domain, various methods~\cite{li2016iterative,zhu20183d,chen20182,zhang2019acfnet,yuan2020object} use multiple stages to refine the segmentation prediction from coarse to fine. ACFNet~\cite{zhang2019acfnet} proposed an attentional class feature module to refine the pixel-wise features based on a coarse segmentation map. OCR~\cite{yuan2020object} further advanced the idea to use a bidirectional connection between pixel-wise features and object-contextual representations to enrich the features. In comparison, refinement modules have been rarely used in point cloud semantic segmentation.

\textbf{Transformer Decoder}
Transformer~\cite{vaswani2017attention} structure has gained huge popularity in recent years. Built on the development of 2D transformer backbones~\cite{dosovitskiy2020vit,Liu2021SwinTH}, various methods~\cite{zhu2021deformable,SETR,wang2021max,cheng2021maskformer,xie2021segformer,zhang2022dino} are proposed to tackle the 2D detection and segmentation problems. Depending on the source of the input, the vision transformers can be categorized into encoder~\cite{Liu2021SwinTH,SETR,xie2021segformer} and decoder~\cite{carion2020end,wang2021max,cheng2021maskformer,yuan2020object,zhang2022dino,li2022mask}. A transformer encoder usually serves as a feature encoding network to replace the conventional neural networks, while a transformer decoder is used to extract class-level or instance-level feature representations for the downstream tasks. In the LiDAR domain, several detection methods~\cite{misra2021end,yang20213d,liu2021,Sheng2021ICCV,bai2022transfusion,nguyen2021boxer,li2022bevformer,Zhou_centerformer} have started to integrate the transformer decoder structure into the previous frameworks. Besides the performance improvement, the transformer decoder demonstrates great potential for an end-to-end training~\cite{misra2021end} and multi-frame~\cite{yang20213d,Zhou_centerformer} / modality~\cite{bai2022transfusion,li2022bevformer} feature fusion. However, studying effective methods of using a transformer decoder in LiDAR segmentation is still an underexplored area. In this paper, we propose a novel class-aware global contextual refinement module for LiDAR segmentation based on the transformer decoder, while exploiting the synergy between detection and segmentation decoders.

\section{Method}

In this section, we present the design of LiDARFormer. As shown in Figure~\ref{fig:architecture}, our framework consists of three parts: (\ref{section:basline}) A 3D encoder-decoder backbone network using 3D sparse convolution; (\ref{section:xsf}) A Cross-space Transformer (XSF) module extracting large-scale and context features in the BEV; (\ref{section:xtf}) A Cross-task Transformer (XTF) decoder that aggregates class-wise and object-wise global contextual information from voxel and BEV feature maps. Our network adopts the multi-task learning framework from LidarMultiNet~\cite{ye2022lidarmultinet}, but further associates the global features between segmentation and detection through a shared cross-task attention layer.

\subsection{Voxel-based LiDAR Perception}
\label{section:basline}
LiDAR point cloud semantic segmentation and object detection aim to predict pixel-wise semantic labels $L=\{l_{i}|l_{i}\in (1\dots K)\}_{i=1}^{N}$ and object bounding boxes $O=\{o_{i}|o_{i}\in\mathbb{R}^{7}\}_{i=1}^{B}$ in a point cloud $P=\{p_{i}|p_{i}\in\mathbb{R}^{3+c}\}_{i=1}^{N}$, where $N$ denotes the number of points, $B$ and $K$ are the number of objects and classes. Each point has $(3+c)$ input features, i.e. the 3D coordinates $(x,y,z)$, the intensity of the reflection, LiDAR elongation, timestamp, etc. Each object is represented by its 3D location, size and orientation. 

\textbf{Voxelization}
We first transform the point cloud coordinates $(x,y,z)$ into the voxel index $\{\mathcal{I}_{i}=(\lfloor \frac{x_{i}}{s_{x}} \rfloor,\lfloor\frac{y_{i}}{s_{y}}\rfloor, \lfloor \frac{z_{i}}{s_{z}}\rfloor)\}_{i=1}^{N}$, where $s$ is the voxel size. Then, we use a simple voxel feature encoder, which only contains a Multi-Layer Perceptron (MLP) and maxpooling layers to generate the sparse voxel feature representation $\mathcal{V}\in\mathbb{R}^{M\times C}$:

\begin{equation}
   \mathcal{V}_{j} = \max\limits_{\mathcal{I}_{i}=\mathcal{I}_{j}}(\textnormal{MLP}(p_{i})) , j\in (1\dots M)
\end{equation}
where $M$ is the number of unique voxel indices. We also generate the ground truth label of each sparse voxel through majority voting: $L_{j}^{v} = \argmax\limits_{\mathcal{I}_{i}=\mathcal{I}_{j}}(l_{i})$.

\textbf{Sparse Voxel-based Backbone Network}
We use a VoxelNet~\cite{zhou2018voxelnet} as the backbone of our network, where the voxel features are gradually downsampled to $\tfrac{1}{8}$ of the original size in the encoder. The sparse voxel features are projected onto the dense BEV map, followed by a 2D multi-scale feature extractor to extract the global information. For the detection task, we attach a detection head to the BEV feature map to predict the object bounding boxes. For the segmentation task, the BEV feature is reprojected to the voxel space, where we use a U-Net decoder to upsample the feature map back to the original scale. We supervise our model with the voxel-level label $L^{v}$ and project the predicted label back to the point level via a de-voxelization step during inference.

\subsection{Cross-space Transformer}
\label{section:xsf}
As shown in Figure~\ref{fig:teaser}, voxel-based LiDAR detection and segmentation generally require the backbone network to extract feature representations on the 2D dense BEV space and 3D sparse voxel space, respectively. To overcome the challenge of merging the features learned from these two tasks, the previous multi-task network~\cite{ye2022lidarmultinet} proposed a global context pooling module to directly map the features based on their location without considering differences in sparsity. In contrast, we propose a cross-space Transformer module that utilizes deformable attention to enhance feature extraction between these spaces to further increase the receptive field.

\begin{figure}[t]
\centering
    \begin{minipage}[b]{1.0\linewidth}
        \includegraphics[width=1.0\linewidth]{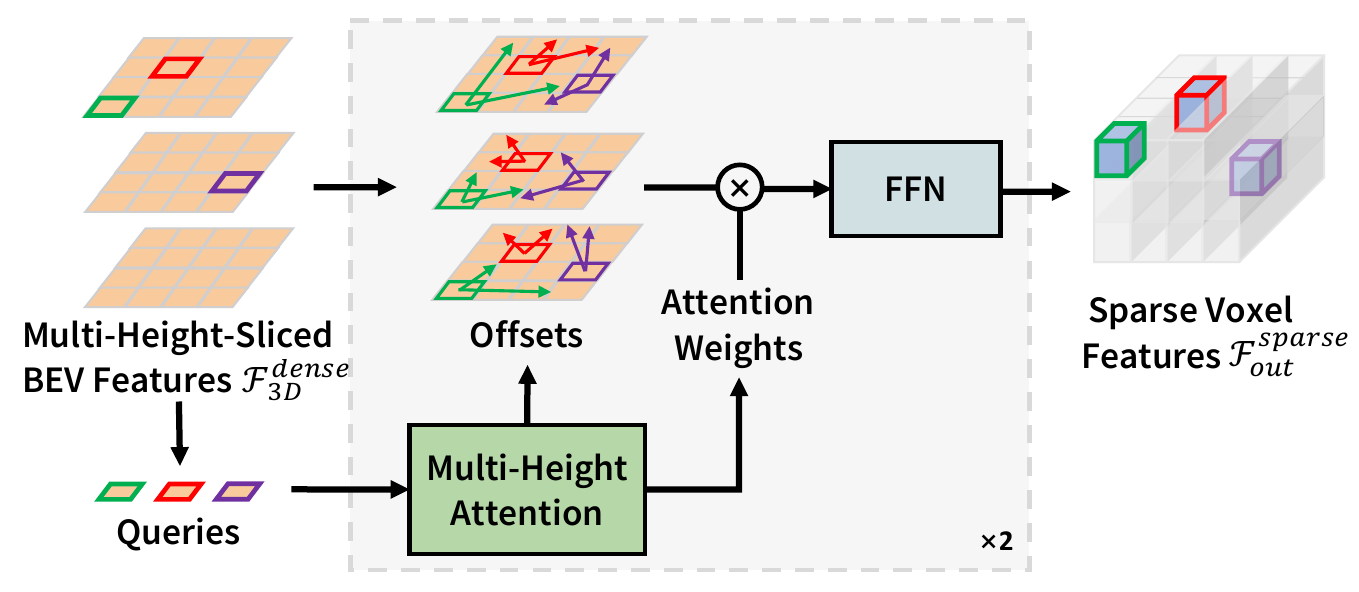}
        \subcaption{Dense-to-sparse Cross-space Transformer}
        \label{fig:xst_a}
    \end{minipage}
    \begin{minipage}[b]{0.9\linewidth}
        \includegraphics[width=1.0\linewidth]{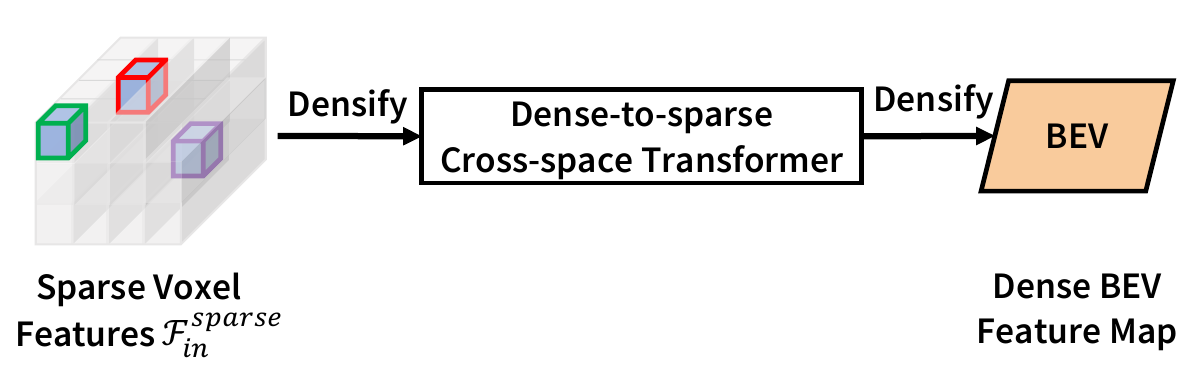} 
        \subcaption{Sparse-to-dense Cross-space Transformer}
        \label{fig:xst_b}
    \end{minipage}
  \caption{\textbf{Illustration of the Cross-space Transformer (XSF) module.} XSF consists of two parts: a multi-height deformable self-attention, and a feed-forward network. (a) convert dense BEV features to sparse voxel features, (b) convert sparse voxel features to dense BEV features with two more densify operations.}
  \label{fig:xst}
\end{figure}

As shown in Figure~\ref{fig:architecture}, we employ a cross-space Transformer to 1) convert the sparse voxel features in the last scale $\mathcal{F}^{sparse}_{in}\in\mathbb{R}^{C\times M'}$ into dense BEV features (\textit{Sparse-to-dense}), and 2) convert the dense BEV features from 2D multi-scale feature extractor $\mathcal{F}^{dense}\in\mathbb{R}^{(C\times \frac{D}{d_{z}}) \times \frac{H}{d_{x}} \times \frac{W}{d_{y}}}$ to sparse voxel features $\mathcal{F}^{sparse}_{out}\in\mathbb{R}^{C\times M'}$, where $d$ is the downsampling ratio and $M'$ is the number of valid voxels in the encoder's last scale (\textit{Dense-to-sparse}). The cross-space Transformer is illustrated in Figure~\ref{fig:xst}. Specifically, in Figure~\ref{fig:xst_a}, $\mathcal{F}^{dense}$ is divided into slices by height as $\mathcal{F}^{dense}_{3D}\in\mathbb{R}^{C\times \frac{D}{d_{z}} \times \frac{H}{d_{x}} \times \frac{W}{d_{y}}}$. Then we take the features from $\mathcal{F}^{dense}_{3D}$ at the valid coordinates $(u, v, h)$ of $\mathcal{F}^{sparse}_{in}$ as query $\textbf{Q}_{3D}$ to predict $\mathcal{F}^{sparse}_{out}$. The deformable attention \cite{zhu2021deformable} is adopted as a self-attention layer to explore global information in the dense feature map. 
Since $\mathcal{F}^{dense}$ lacks height information, due to the fact that 2D multi-scale feature extractor mainly focuses on BEV-level information,
we develop a multi-head multi-height attention module to learn features along all heights: For every reference voxel whose location is $\xi = (u, v)$ on the sliced BEV feature map at height $h$, the deformable self-attention uses a linear layer to learn BEV offsets $\Delta \xi$ at all heads and heights.
The features at $\xi + \Delta \xi$ will be sampled from different multi-heights-sliced BEV feature maps through bilinear interpolation. 
The output of the multi-height deformable self-attention $\chi (p)$ can be formulated as:
\begin{equation}
    \chi (p) = \sum_{i=1}^{N_{head}}{W_{i}[\sum_{j=1}^{N_{height}} \sum_{r=1}^{R} \sigma (W_{ijr} q_{p}) W_{i}^{'} x^{j} (\xi + \Delta \xi_{ijr})]}
\end{equation}
where $N_{head}$ is the number of heads, $N_{height} = \frac{D}{d_{z}}$ is the number of heights, $W$ is learnable weights, $R$ is the number of sampling points, $x^{j}$ is the multi-heights-sliced BEV features, $q_{p}$ is the query features at the position $\xi$, and $\sigma (W_{ijr} q_{p})$ is the attention weight.

Since the Dense-to-sparse cross-space Transformer is applied after the 2D feature extractor, it will not affect the learned 2D BEV features, thus has limited impact on increasing the detection performance. To  increase the receptive field of the 2D BEV feature extractor, we add a cross-space Transformer module converting $\mathcal{F}^{sparse}_{in}$ into dense BEV features in a similar manner, as shown in Figure~\ref{fig:xst_b}. It equips the BEV feature which will be fed into a 2D multi-scale feature extractor with more context information.  

\subsection{Cross-task Transformer Decoder}
\label{section:xtf}

Although object detection and semantic segmentation share correlated information, they are usually learned in two separate network structures. LidarMultiNet~\cite{ye2022lidarmultinet} demonstrates that through sharing intermediate feature representation, both detection and segmentation performance can get improved. However, no high-level information is shared during the training of the multi-task network. To further explore the multi-task learning synergy, we propose to use a shared transformer decoder to bridge between the class-level information from segmentation and the object-level information from detection. In this section, we first present a novel segmentation decoder that uses class feature embedding to perform dynamic segmentation. Then, we introduce an approach to connect this segmentation decoder with the conventional detection decoder through cross-task attention.

\textbf{Segmentation Transformer Decoder} Inspired by the coarse-to-fine methods~\cite{zhang2019acfnet, yuan2020object} in the 2D image segmentation, we propose a class-aware feature refinement module to enhance the global information learning for the segmentation task. We use an initial segmentation prediction to generate the class feature embedding. Then, we use a transformer with bidirectional cross-attention to refine both voxel and class feature representations. The class feature representation is also served as the dynamic kernel in the later segmentation head.

Given an initial semantic segmentation score $y=\{pred_{j}|pred_{j}\in [0,1]^{K}\}_{j=1}^{M}$, and its encoded feature representation $\mathcal{F} \in \mathbb{R}^{M\times C}$, where $M$ is the number of valid predictions, we generate the class feature embedding $\varepsilon = \{\varepsilon_{k}|k\in\{1\dots K\}\}$ as follows: 
$\varepsilon_{k} = \frac{\sum_{j=1}^{M}pred_{j}[k]\cdot \mathcal{F}_{j}}{\sum_{j=1}^{M}pred_{j}[k]}$.
In our cross-task transformer, we use a coarse prediction and its corresponding BEV features to initialize the class feature embedding.
The class feature embedding $\varepsilon$ encapsulates the class center information based on the coarse segmentation result of each scan. Assuming that points from the same class have similar or correlated features in the encoded feature embedding, the learned class features can help the network distinguish the edge points that are ambiguous in the segmentation head.

\begin{figure}[t]
  \begin{center}
    \includegraphics[width=1\linewidth]{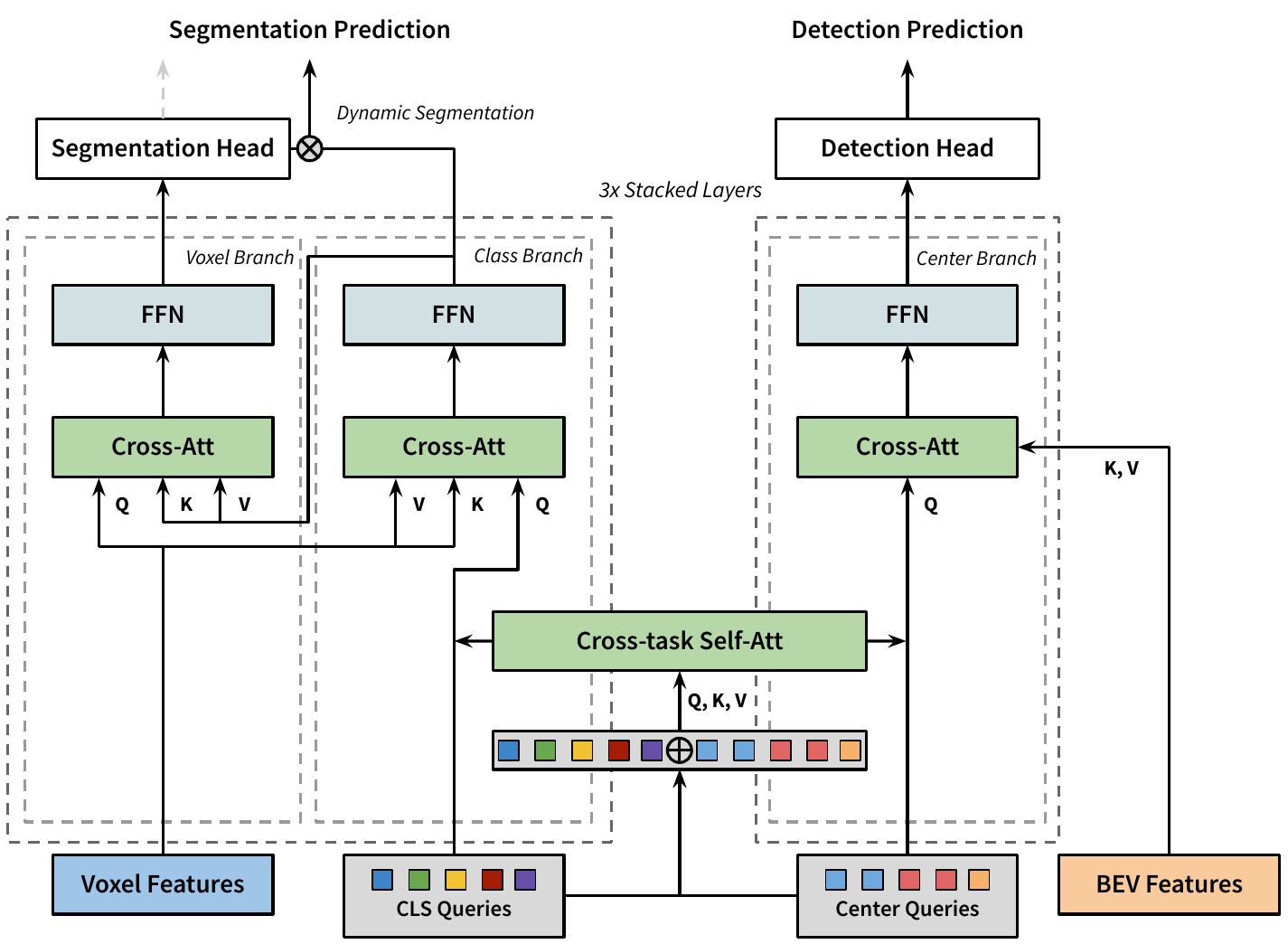}
  \end{center}
  \caption{\textbf{Cross-task Transformer (XTF).} The segmentation and detection decoders share a self-attention layer to transfer the cross-task features. In the segmentation decoder, we use a bidirectional cross-attention to refine voxel features based on the aggregated class feature embedding. For simplicity, the skip connection and the layer norm are ignored in this figure.}
  \label{fig:transformer}
  \vspace{-10pt}
\end{figure}

Similar to ~\cite{yuan2020object}, we propose to use a transformer decoder to further extract the class feature embedding and refine the original voxel features simultaneously through bidirectional cross-attention. As shown in Figure~\ref{fig:transformer}, our transformer structure has two parallel branches for the voxel feature $\mathcal{V} \in \mathbb{R}^{M\times C}$ and the class feature $\varepsilon \in \mathbb{R}^{K\times C}$.

We use a standard transformer decoder~\cite{vaswani2017attention}, containing a multi-head self-attention layer, a multi-head cross-attention layer, and a feed-forward layer, to extract class features using  $\varepsilon$ as the initial query embedding. In the cross-attention layer, query $\textbf{Q}_{c}$ is the linear projection of $\varepsilon$, while key $\textbf{K}_{v}$ and value $\textbf{V}_{v}$ are the linear projection of $\mathcal{V}$. It can be formulated as :

\begin{equation}
    \textnormal{CrossAtt}(\mathcal{V}\to \varepsilon) = \textnormal{Softmax}(\frac{\textbf{Q}_{c}\textbf{K}_{v}^{T}}{\sqrt{C}})\textbf{V}_{v}.
\end{equation}

Next, we use an inverse transformer decoder to transfer the encoded class features back to the voxel features. It is infeasible to use self-attention in the voxel branch due to the huge size of the voxels. Conversely, query $\textbf{Q}_{v}$ is from the linear projection of $\mathcal{V}$, key $\textbf{K}_{c}$, and value $\textbf{V}_{c}$ are the linear projection of the output $\varepsilon'$ in the class branch:

\begin{equation}
    \textnormal{CrossAtt}(\varepsilon' \to \mathcal{V}) = \textnormal{Softmax}(\frac{\textbf{Q}_{v}\textbf{K}_{c}^{T}}{\sqrt{C}})\textbf{V}_{c}.
\end{equation}

The output voxel feature $\mathcal{V}'$ is then concatenated to the original features $\mathcal{V}^{r} = (\mathcal{V}, \mathcal{V}')$ for the segmentation head.

\textbf{Dynamic Kernel}
Conventional segmentation networks use a segmentation head that consists of convolution or linear layers to reduce the channel size of a voxel feature to the number of classes to make the prediction. The weights learned in the segmentation head are shared among different frames. Therefore the segmentation head is hard to adjust to the varying conditions of scenes. Following the new trend in the image instance segmentation~\cite{wang2020solov2,wang2021max,cheng2021maskformer,li2022mask}, we directly use the learned class feature embedding $\varepsilon'$ as the kernel to generate the semantic logits $\mathcal{S} = \frac{\Phi(\mathcal{V}^{r} ) \cdot \varepsilon'^{T}}{\sqrt{C}} \in \mathbb{R}^{M\times K}$, where $\Phi$ is the convolution layer that reduces the channel size of the voxel feature to $C$. 

\begin{table*}[t]
\begin{minipage}{.34\textwidth}
\centering
\caption{Detection results on the \texttt{test} split of nuScenes. ``TTA" means test-time augmentation. }
\label{tab:nusc_det_test}
\resizebox{\linewidth}{!}{%
\begin{tabular}{l|c|c>{\columncolor[gray]{0.95}}c}
\Xhline{4\arrayrulewidth}
    Model & Ref & mAP & NDS \\
    \Xhline{2\arrayrulewidth}
    CBGS~\cite{zhu2019class} & arXiv 2019 & 52.8 & 63.3 \\
    CenterPoint~\cite{yin2021center} & CVPR 2021 & 58.0 & 65.5 \\
    HotSpotNet~\cite{chen2020object} & ECCV 2020 & 59.3 & 66.0 \\
    Object DGCNN~\cite{wang2021object} & NeurIPS 2021 & 58.7 & 66.1 \\
    AFDetV2~\cite{hu2022afdetv2} & AAAI 2022 & 62.4 & 68.5 \\
    Focals Conv~\cite{Chen2022focalsparse} & CVPR 2022 & 63.8 & 70.0 \\
    TransFusion-L~\cite{bai2022transfusion} & CVPR 2022 & 65.5 & 70.2\\
    LargeKernel3D~\cite{chen2022scaling}  & CVPR 2023 & 65.3 & 70.5 \\
    SphereFormer~\cite{ye2022lidarmultinet}  & CVPR 2023 & 65.5 & 70.7 \\
    LidarMultiNet~\cite{ye2022lidarmultinet}  & AAAI 2023 & 67.0 & 71.6 \\
    MDRNet-TTA~\cite{huang2022rethinking} & arXiv 2022 & 67.2 & 72.0 \\
    LargeKernel3D-TTA~\cite{chen2022scaling}  & CVPR 2023 & 68.8 & 72.8 \\
    FocalFormer3D-TTA~\cite{chen2023focalformer3d}  & ICCV 2023 & 70.5 & 73.9 \\
    \Xhline{2\arrayrulewidth}
    LiDARFormer    && 68.9 & 72.4  \\
    LiDARFormer-TTA    && \textbf{71.5} & \textbf{74.3}  \\
    
    \Xhline{4\arrayrulewidth}
\end{tabular}
}
\end{minipage}
\hfill
\begin{minipage}{.3\textwidth}
\centering
\caption{Segmentation results on the \texttt{test} split of nuScenes.}
\label{tab:nusc_seg_test}
\resizebox{\linewidth}{!}{%
\begin{tabular}{l|c|>{\columncolor[gray]{0.95}}c}
    \Xhline{4\arrayrulewidth}
    Model & Ref & mIoU \\
    \Xhline{2\arrayrulewidth}
    PolarNet~\cite{Zhang_2020_CVPR} & CVPR 2020 &  69.8\\
    PolarStream~\cite{chen2021polarstream} & NeurIPS 2021 &  73.4\\
    JS3C-Net~\cite{yan2020sparse} & AAAI 2021 &  73.6 \\
    Cylinder3D~\cite{zhu2021cylindrical} & CVPR 2021 &  77.2 \\
    AMVNet~\cite{liong2020amvnet} & arXiv 2020&  77.3 \\
    SPVNAS~\cite{tang2020searching} & ECCV 2020 &  77.4 \\
    Cylinder3D++~\cite{zhu2021cylindrical} & CVPR 2021 &  77.9 \\
    AF2S3Net~\cite{Cheng_2021_CVPR} & CVPR 2021 &  78.3 \\
    GASN~\cite{ye2022efficient} & ECCV 2022 &  80.4 \\
    SPVCNN++~\cite{tang2020searching} & ECCV 2020 &81.1 \\
    LidarMultiNet~\cite{ye2022lidarmultinet} & AAAI 2023 & 81.4 \\
    \Xhline{2\arrayrulewidth}
    
    LiDARFormer &&  81.0 \\
    LiDARFormer-TTA &&  \textbf{81.5} \\
    
    \Xhline{4\arrayrulewidth}
\end{tabular}
}
\end{minipage}
\hfill
\begin{minipage}{.31\textwidth}
\centering
    \caption{Results on the \texttt{val} split of nuScenes. 
    *: Reported by ~\cite{zhu2021cylindrical}.}
    \label{tab:nusc_val}
    \resizebox{\linewidth}{!}{%
    \begin{tabular}{l|>{\columncolor[gray]{0.95}}c|c>{\columncolor[gray]{0.95}}c}
    \Xhline{4\arrayrulewidth}
    Model & mIoU & mAP & NDS \\
    \Xhline{2\arrayrulewidth}
    RangeNet++~\cite{milioto2019rangenet++} & 65.5* & - & -  \\
    PolarNet~\cite{Zhang_2020_CVPR}   & 71.0*     & - & -       \\
    SalsaNext~\cite{cortinhal2020salsanext}  & 72.2*     & - & -        \\
    AMVNet~\cite{liong2020amvnet}     & 77.2     & - & -        \\
    Cylinder3D~\cite{zhu2021cylindrical} & 76.1     & - & -       \\
    RPVNet~\cite{xu2021rpvnet}     & 77.6     & - & -         \\
    SphereFormer~\cite{lai2023spherical}     & 78.4     &   &           \\
    \Xhline{2\arrayrulewidth}
    CBGS~\cite{zhu2019class} & - & 51.4 & 62.6\\
    CenterPoint~\cite{yin2021center} & - & 57.4 & 65.2\\
    TransFusion-L~\cite{bai2022transfusion} & - & 60.0 & 66.8 \\
    BEVFusion-L~\cite{liu2022bevfusion} & - & 64.7 & 69.3 \\
    \Xhline{2\arrayrulewidth}
    LidarMultiNet~\cite{ye2022lidarmultinet} & 82.0 & 63.8 & 69.5 \\
    \Xhline{2\arrayrulewidth}
    LiDARFormer seg only    & 81.7    & - & -   \\
    LiDARFormer     & \textbf{82.7}    & \textbf{66.6} & \textbf{70.8}    \\
    \Xhline{4\arrayrulewidth}
    \end{tabular}
    }
\end{minipage}

\end{table*}

\textbf{Cross-task Attention}
As shown in Figure~\ref{fig:transformer}, we adopt the detection transformer decoder from the well-studied CenterFormer~\cite{Zhou_centerformer}, which represents the object-level feature as center query embedding initialized from BEV center proposals. We initialize the class feature embedding $\varepsilon$ using the BEV feature.
Class and center features are concatenated and then sent into a shared transformer decoder, where the information between detection and segmentation tasks are transferred to each other through a cross-task self-attention layer. Due to the memory limitation, the class and center feature aggregate features separately from the voxel and BEV feature maps, respectively.

\section{Experiments}
In this section, we present the experimental results of our proposed method on two large-scale public LiDAR point cloud datasets: the nuScenes dataset~\cite{caesar2020nuscenes} and the Waymo Open Dataset\cite{sun2020scalability}, both of which have 3D object bounding boxes and pixel-wise semantic label annotations. We also provide a detailed ablation study of the improvements and in-depth analysis of our model. More details and visualization are included in the supplementary materials.

\subsection{Datasets}
The \textbf{NuScenes} dataset is a large-scale autonomous driving dataset developed by Motional. It contains 1000 scenes of 20s video data, each of them captured by a 20Hz Velodyne HDL-32E Lidar sensor with 32 vertical beams. NuScenes provides annotations of object bounding boxes and pixel-wise semantic labels at each keyframe sampled at 2 Hz. 16 classes are used for the semantic segmentation evaluation. 10 foreground object (``thing'') classes with ground truth bounding box labels are used for the object detection task. For the nuScenes detection task, mean Average Precision (mAP) and NuScenes Detection Score (NDS) are used as the metrics. For semantic segmentation, mean Intersection over Union (mIoU) is used as the metric.

The \textbf{Waymo Open Dataset (WOD)} contains around 2000 scenes of 20s video data that is collected at 10Hz by a 64-line LiDAR sensor. Even though WOD provides object bounding box annotation for every frame, it only has the semantic annotation in some key frames sampled at 2Hz. WOD has semantic labels for 23 classes and uses the standard mIoU as the evaluation metric. For the object bounding box annotation, 3 classes of vehicles, pedestrians, and cyclists are calculated into the 3D detection metrics. Average Precision Weighted by Heading (APH) is used as the main detection evaluation metric. The ground truth objects are categorized into two levels of difficulty, LEVEL\_1 (L1) is assigned to the examples that have more than 5 LiDAR points and not in the L2 category, while LEVEL\_2 (L2) is assigned to examples that have at least 1 LiDAR point and at most 5 points or are manually labeled as hard. The primary metric mAPH L2 is computed by considering both L1 and L2 examples.

\subsection{Experiment Setup}
We used the AdamW optimizer with the one-cycle scheduler to train our model for 20 epochs. Most experiments are conducted on 8 Nvidia A100 GPUs with batch size 16. For the multi-task training experiments on WOD, we used batch size 8 because of the GPU memory limits. We used the voxel size of $[0.1,0.1,0.2]$ for nuScenes datasets, and $[0.1,0.1,0.15]$ for Waymo Open Dataset. For the segmentation task, we used a combination of cross-entropy loss and Lovasz loss~\cite{berman2018lovasz} to optimize our network. For the detection task, we followed \cite{yin2021center} to use the common center heatmap classification loss and bounding box regression loss. We added an auxiliary loss on the output voxel features or BEV features to supervise the segmentation prediction, which is used to initialize the class feature embedding. All losses are fused by multi-task uncertainty weighting strategy~\cite{kendall2018multi}. We concatenated the points from the previous 9 scans to the current point cloud in nuScenes, and 2 scans in WOD. Standard data augmentation strategy~\cite{wang2021pointaugmenting,ye2022lidarmultinet} were applied when training the model. More network and training details are included in the supplementary materials.

\begin{table*}[t]
\begin{minipage}{.56\textwidth}
\centering
\caption{Detection L2 mAPH results on the \texttt{test} split of WOD. ``L" and ``CL" denote LiDAR-only and camera \& LiDAR fusion methods. Second best results are underlined.}
\label{tab:WOD_det_test}
\resizebox{\linewidth}{!}{
\begin{tabular}{l|c|c|c|ccc|>{\columncolor[gray]{0.95}}c}
\Xhline{4\arrayrulewidth}
    Model & Ref & Modal & Frame & Veh. & Ped. & Cyc. & Mean \\
    \Xhline{2\arrayrulewidth}
    M3DETR~\cite{guan2022m3detr} & WACV 2022 & L & 1 & 70.0  & 52.0 & 63.8 & 61.9\\
    PV-RCNN++~\cite{shi2021pv} & arXiv 2022 & L & 1 & 73.5 & 69.0 & 68.2 & 70.2\\
    CenterPoint++~\cite{yin2021center} & CVPR 2021 & L & 3 & 75.1 & 72.4 & 71.0 & 72.8\\
    SST\_3f~\cite{fan2022embracing} & CVPR 2022 & L & 3 & 72.7 & 73.5 & 72.2 & 72.8\\
    AFDetV2~\cite{hu2022afdetv2} & AAAI 2022 & L & 2 & 73.9  & 72.4 & 73.0 & 73.1 \\
    DeepFusion \cite{li2022deepfusion} & CVPR 2022 & CL & 5 &  75.7 & 76.4 & 74.5 & 75.5  \\
    MPPNet~\cite{chen2022mppnet} & ECCV 2022 & L & 16 & 76.9  & 75.9 & 74.2 & 75.7 \\
    CenterFormer~\cite{Zhou_centerformer} & ECCV 2022 & L & 16 & \textbf{78.3}  & \textbf{77.4} & 73.2 & \underline{76.3} \\
    BEVFusion~\cite{liu2022bevfusion} & ICRA 2023 & CL  & 3 & \underline{77.5}  & 76.4 & \textbf{75.1} & \underline{76.3}  \\
    \Xhline{2\arrayrulewidth}
    LiDARFormer    & & L & 3  & \underline{77.5} & \underline{77.2} & \underline{74.6} & \textbf{76.4}\\
    \Xhline{4\arrayrulewidth}
\end{tabular}
}
\end{minipage}
\hfill
\begin{minipage}{.42\textwidth}
\centering
	\caption{Results on \texttt{val} split of WOD. *: From our reproduction.}
	\label{tab:waymo_val}
	\resizebox{\linewidth}{!}{%
	\begin{tabular}{l|c|c|>{\columncolor[gray]{0.95}}c>{\columncolor[gray]{0.95}}c}
		\Xhline{4\arrayrulewidth}
		Model & Ref & Frame & mIoU & L2 mAPH \\
		\Xhline{2\arrayrulewidth}
		PolarNet~\cite{Zhang_2020_CVPR} & CVPR 2020 & 1 &  61.6* & - \\
		Cylinder3D~\cite{zhu2021cylindrical} & CVPR 2021 & 1 &  66.6* & - \\
        SphereFormer~\cite{lai2023spherical} & CVPR 2023 & - &  69.9 & - \\
		\Xhline{2\arrayrulewidth}
		PV-RCNN++~\cite{shi2021pv} & IJCV 2022 & 1 & - & 68.6 \\
		AFDetV2-Lite~\cite{hu2022afdetv2}& AAAI 2022 & 1 & - & 68.8 \\
		CenterPoint++~\cite{yin2021center}& CVPR 2021 & 3 & - & 71.6 \\
        FlatFormer~\cite{liu2023flatformer}& CVPR 2023 & 3 & - & 72.0 \\
		SST~\cite{fan2022embracing} & CVPR 2022 & 3 & - & 72.4 \\
        DSVT~\cite{wang2023dsvt}& CVPR 2023 & 3 & - & 75.5 \\
		CenterFormer~\cite{Zhou_centerformer}& ECCV 2022 & 8 & - & 73.7 \\
		MPPNet~\cite{chen2022mppnet}& ECCV 2022 & 16 & - &  74.9 \\
		\Xhline{2\arrayrulewidth}
		LidarMultiNet~\cite{ye2022lidarmultinet}& AAAI 2023& 3 & 71.9 & 75.2\\
		\Xhline{2\arrayrulewidth}
		LiDARFormer seg only & - & 3 &  71.3 & - \\
		LiDARFormer  & - & 3 &  \textbf{72.2} & \textbf{76.2}\\
		
		\Xhline{4\arrayrulewidth}
	\end{tabular}
	}
\end{minipage}
\end{table*}

\begin{table*}[t]
\begin{minipage}{.49\textwidth}
\centering
\caption{The ablation of mIoU improvement of each component on the nuScenes and WOD \texttt{val} split when trained only for the segmentation task. XSF and STD stand for cross-space transformer and segmentation transformer decoder.}
\label{table:ablation_seg_only}
\resizebox{\linewidth}{!}{
\begin{tabular}{c|ccc|cc}
\Xhline{4\arrayrulewidth}
Baseline (\ref{section:basline}) & STD & Multi-frame & XSF & nuScenes & WOD \\ \Xhline{2\arrayrulewidth}
$\checkmark$ &              &              &              & 76.6\phantom{ (+1.7)}   &  70.3\phantom{ (+1.7)} \\
$\checkmark$ & $\checkmark$ &              &              & 78.3 (\textcolor{red}{+1.7})   &  70.6 (\textcolor{red}{+0.3}) \\
$\checkmark$ & $\checkmark$ & $\checkmark$ &              & 80.8 (\textcolor{red}{+4.2})   &  71.2 (\textcolor{red}{+0.9}) \\
$\checkmark$ & $\checkmark$ & $\checkmark$ & $\checkmark$ & \textbf{81.7} (\textcolor{red}{+5.1})   &  \textbf{71.3} (\textcolor{red}{+1.0})
 \\\Xhline{4\arrayrulewidth}
\end{tabular}
}
\end{minipage}
\hfill
\begin{minipage}{.49\textwidth}
\centering
\caption{The ablation of the improvement of shared transformer decoder on the nuScenes \texttt{val} split when jointly trained with detection task.}
\label{table:ablation}
\resizebox{\linewidth}{!}{
\begin{tabular}{c|ccc|>{\columncolor[gray]{0.95}}cc>{\columncolor[gray]{0.95}}c}
\Xhline{4\arrayrulewidth}
& \multicolumn{2}{c}{XTF} & & & & \\
\multirow{-2}{*}{Baseline~\cite{ye2022lidarmultinet}} &{Seg} &{Det} & \multirow{-2}{*}{XSF} & \multirow{-2}{*}{mIoU} & \multirow{-2}{*}{mAP}  & \multirow{-2}{*}{NDS}\\ \Xhline{2\arrayrulewidth}
$\checkmark$ &              &              &              & 81.8\phantom{ (+1.7)}  &  65.2\phantom{ (+1.7)} & 70.0\phantom{ (+1.7)} \\
$\checkmark$ & $\checkmark$ &              &              & 82.1 (\textcolor{red}{+0.3})  &  65.4 (\textcolor{red}{+0.2}) & 70.2 (\textcolor{red}{+0.2}) \\
$\checkmark$ &              & $\checkmark$ &              & 82.4 (\textcolor{red}{+0.6})  &  65.9 (\textcolor{red}{+0.7}) & 70.3 (\textcolor{red}{+0.3})\\
$\checkmark$ & $\checkmark$ & $\checkmark$ &              & 82.6 (\textcolor{red}{+0.8})   &  66.0 (\textcolor{red}{+0.8}) & 70.2 (\textcolor{red}{+0.2}) \\
$\checkmark$ & $\checkmark$ & $\checkmark$ & $\checkmark$ & \textbf{82.7} (\textcolor{red}{+0.9})  & \textbf{66.6} (\textcolor{red}{+1.4}) & \textbf{70.8} (\textcolor{red}{+0.8})\\\Xhline{4\arrayrulewidth}
\end{tabular}
}
\end{minipage}
\end{table*}

\subsection{Main Results}
We present the detection and segmentation benchmark results on both nuScenes and WOD. All results of other methods in the test set are from the literature, where most of them apply test-time augmentation (TTA) or an ensemble method to increase the performance. In addition to our multi-task network, we also provide the results of the segmentation-only variation of our model, which is trained only with the segmentation transformer decoder.

\textbf{NuScenes} In Table~\ref{tab:nusc_det_test} and Table~\ref{tab:nusc_seg_test}, we compare LiDARFormer with other state-of-the-art methods on the test set of nuScenes. LiDARFormer reaches the top performance of $81.5\%$ mIoU, $71.5\%$ mAP, and $74.3\%$ NDS for a single model result. Notably, the results of the detection task outperform all previous methods by a large margin, especially for the mAP metric. Although the segmentation performance of LiDARFormer is only $0.1\%$ higher than LidarMultiNet, LiDARFormer does not require a second stage and can be trained end-to-end by comparison. To fairly compare with other methods without the effect of test-time augmentation, we also demonstrate the performance on the validation set of nuScenes in Table~\ref{tab:nusc_val}. Our segmentation-only LiDARFormer achieves a $81.7\%$ mIoU performance while full LiDARFormer further improves the mIoU to $82.7\%$ with the SOTA detection performance NDS $70.8\%$. Our method surpasses all previous state-of-the-art methods, which matches our result in the test set.

\textbf{Waymo Open Dataset} Table~\ref{tab:WOD_det_test} shows the detection results of LiDARFormer on the test set of WOD. LiDARFormer achieves the state-of-the-art performance of $76.4\%$ L2 mAPH, outperforming even the camera-LiDAR fusion methods and methods that use a much greater number of frames. Lastly, we report the validation results on Waymo Open Dataset in Table~\ref{tab:waymo_val}. We reproduce the result of PolarNet and Cylinder3D based on their released code for comparison. Our segmentation-only LiDARFormer achieves a $71.3\%$ mIoU performance on the validation set. Our multi-task model also outperforms the previous best multi-task network by $0.3\%$ on the segmentation task. For the more competitive detection task, our method reaches the best L2 mAPH result of $76.2\%$.

\subsection{Ablation Study}

\textbf{Effect of Transformer Structure on Segmentation Task}  Table~\ref{table:ablation_seg_only} shows the effectiveness of each proposed component in our method when trained only for the segmentation task. We use the network described in \ref{section:basline} as our baseline model. This simple design already can achieve competitive performance compared to other current state-of-the-art methods. After adding the segmentation transformer decoder, the mIoU increases by $1.7\%$ and $0.3\%$ in nuScenes and WOD, respectively. By concatenating points from previous frames to the current frame, the result further increases by $2.5\%$ and $0.6\%$. The cross-space transformer also can improve the mIoU by $0.9\%$ and $0.1\%$, respectively.

\textbf{Effect of the Unified Multi-task Transformer Decoder}
Table~\ref{table:ablation} demonstrates the improvements achieved by our proposed transformer decoder in the multi-task network. We use the 1st-stage results of LidarMultiNet~\cite{ye2022lidarmultinet} as our baseline. Adding an individual transformer decoder to either the detection or segmentation branch results in improved performance in both tasks, as our multi-task network has a shared backbone, allowing improvement in one task to contribute to feature representation learning. Our proposed shared transformer decoder yields superior overall performance by introducing cross-task attention learning. The cross-space transformer module further improves performance, particularly for the detection task. We also evaluate the panoptic segmentation performance of our multi-task network in Table~\ref{table:pano}. Even without a second stage dedicated to panoptic segmentation, our model achieves competitive results compared to the previous best method, LidarMultiNet. This demonstrates the ability of our multi-task transformer decoder to generate more compatible results for both tasks.
\begin{table}[t]
    \centering
    \caption{Panoptic segmentation result on nuScenes \texttt{val} split.}
    \label{table:pano}
    \resizebox{1.0\linewidth}{!}{
    \begin{tabular}{l|c|cccc}
    \Xhline{4\arrayrulewidth}
         & stage & PQ & SQ & RQ & mIoU\\ 
        \Xhline{2\arrayrulewidth}
        LidarMultiNet~\cite{ye2022lidarmultinet} & 2-stage & \textbf{81.8} & \textbf{90.8} & 89.7 & 83.6 \\
        LiDARFormer & 1-stage & \textbf{81.8} & 90.7 & \textbf{89.9} & \textbf{84.1} \\
    \Xhline{4\arrayrulewidth}
    \end{tabular}
    }
\end{table}

\subsection{Analysis}

\textbf{Analysis of the Segmentation Decoder} We compare the segmentation-only performance of our method using different transformer designs in Table~\ref{table:ccr}. Removing either way of the cross-attention leads to an inferior result. The dynamic kernel design outperforms the traditional segmentation head by $0.8\%$. Furthermore, the performance is $0.3\%$ lower without using an auxiliary segmentation head to initialize the class embedding.

\begin{table}[t]
\centering
    \caption{Design choice of the segmentation decoder on the nuScenes \texttt{val} split.}

    \label{table:ccr}
    \resizebox{\linewidth}{!}{%
    \begin{tabular}{l|l}
    \Xhline{4\arrayrulewidth}
    LiDARFormer seg only result without XSF (mIoU) & 80.8 \\
    \Xhline{2\arrayrulewidth}
    w/o voxel to class attention & 80.4 (\textcolor{green}{-0.4}) \\
    w/o class to voxel attention  & 80.1 (\textcolor{green}{-0.7}) \\
    w/o dynamic kernel  &  80.3 (\textcolor{green}{-0.5})   \\
    w/o class embedding initialization & 80.5 (\textcolor{green}{-0.3}) \\
    \Xhline{4\arrayrulewidth}
    \end{tabular}
    }

\end{table}

\begin{table}[t]
\centering
    \caption{The ablation of XSF on the nuScenes \texttt{val} split. S$\rightarrow$D and D$\rightarrow$S denote sparse-to-dense (\ref{fig:xst_b}) and dense-to-sparse (\ref{fig:xst_a}) XSFs.}
    \label{table:xsf}
    \resizebox{\linewidth}{!}{%
    \begin{tabular}{ccc|ccc}
    \Xhline{4\arrayrulewidth}
    S$\rightarrow$D & D$\rightarrow$S & Add Convs & mIoU & mAP & NDS \\
    \Xhline{2\arrayrulewidth}
    \multicolumn{6}{c}{Segmentation Only}\\
    \Xhline{2\arrayrulewidth}
    $\checkmark$ & $\checkmark$ & & 81.7\phantom{ (-1.7)} & - & - \\
     & & $\checkmark$ &  80.9 (\textcolor{green}{-0.8}) & - & - \\
    \Xhline{2\arrayrulewidth}
    \multicolumn{6}{c}{Multi-task}\\
    \Xhline{2\arrayrulewidth}
    $\checkmark$ & $\checkmark$ & & 82.7\phantom{ (+1.7)}  & 66.6\phantom{ (-1.7)} & 70.8\phantom{ (-1.7)}\\
     & $\checkmark$ & $\checkmark$ & 82.8 (\textcolor{red}{+0.1}) & 66.0 (\textcolor{green}{-0.6}) & 70.5 (\textcolor{green}{-0.3}) \\
    \Xhline{4\arrayrulewidth}
    
    \end{tabular}
    }
\end{table}

\textbf{Analysis of Cross-space Transformer} Table~\ref{table:xsf} illustrates the effectiveness of the Cross-Space Transformer (XSF) module in both detection and segmentation tasks, as compared to the direct mapping method. If we replace XSF with additional convolution layers of similar parameter size, the segmentation performance decreases by 0.8\%. However, when we only replace the sparse-to-dense XSF in the multi-task model, the segmentation performance remains largely unaffected, while detection performance shows a significant decline. This finding suggests that the dense-to-sparse and sparse-to-dense XSFs contribute differently to the detection and segmentation tasks.

\begin{figure}[t]
  \begin{center}
    \includegraphics[width=1\linewidth]{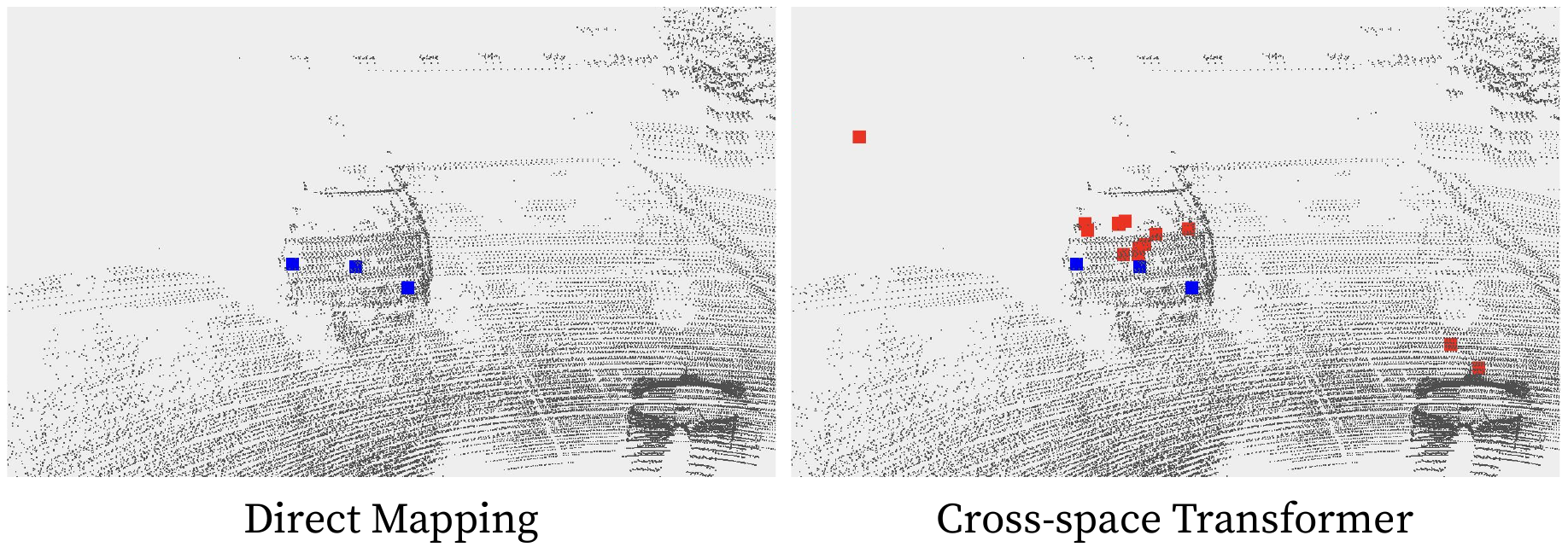}
  \end{center}
  \caption{\textbf{Visualization of the learned offsets.} We showcase the features of a car's 3D voxels (\textcolor{blue}{blue}) and their corresponding deformable offsets (\textcolor{red}{red}) that were learned in our XSF module. For a better visual representation, we only highlight the offsets with high attention scores.}
  \label{fig:offsets}
\end{figure}

In Figure~\ref{fig:offsets}, we provide a visualization of the deformable offsets in our cross-space transformer. When using the previous direct mapping method for sparse voxels, only the features in the same position are used for transferring features between 3D and 2D space. This method may not utilize some useful features learned in the dense 2D BEV map. In contrast, our method is capable of aggregating related features across a wider range.

\section{Conclusion}

In this paper, we present \textbf{LiDARFormer}, a novel and effective paradigm for multi-task LiDAR perception. 
Our method offers a novel way of strengthening voxel feature representation and enables joint learning of detection and segmentation tasks in a more elegant and effective manner. 
Although we have designed LiDARFormer for LiDAR-only input, our transformer XSF and XTF can extend to learn multi-modality and temporal features simply through cross-attention layers. Similarly, XSF can apply multi-scale feature maps in the deformable attention module to further extract the contextual information with larger receptive fields.
LiDARFormer sets a new state-of-the-art performance on the competitive nuScenes and Waymo detection and segmentation benchmarks. We believe that our work will inspire more innovative future research in this field. 

\appendix
\section{Network Details}
\textbf{Voxel Feature Encoder} We adopt the same design as ~\cite{Zhang_2020_CVPR} to encode the point cloud into a voxel feature map. First, we group points within each voxel together and append 6 additional features to the point features $P\in \mathbb{R}^{3+c+6}$, i.e. the center of corresponding voxel $(x_{v},y_{v},z_{v})$ and the offset to the center $(x-x_{v},y-y_{v},z-z_{v})$. Next, we use 4 stacked layers of MLP to transform the point feature to a high dimensional space, followed by a sparse max pooling layer to extract voxel feature representation in each valid voxel. The channel size is $[64,128,256,256]$ in each MLP.

\textbf{XSF structure} We apply 2 stacked transformer blocks, each with 4 heads of deformable self-attention. We use a channel size of 64 in each head of Dense-to-Sparse XSF and a channel size of 32 in each head of Sparse-to-Dense XSF. The channel size of the FFN is 256 in both XSFs. We use pre-norm rather than post-norm in each layer.

\textbf{XTF structure} We use 3 stacked transformer decoder layers, each with 4 heads of self-attention and cross-attention. We use a channel size of 32 in each head and channel size of 64 in the FFN.

\section{More Discussions}
\textbf{More Analysis of XSF} 
In Table~\ref{table:query}, we show the results of adopting different types of query in the dense-to-sparse XSF. The features in the BEV feature map at valid voxels serve as queries in our LiDARFormer. Voxel query refers to taking the features from a sparse feature map at valid voxels while embedding query means treating the embeddings of the valid coordinates as queries. The performance slightly drops by $0.3\%$ and $0.4\%$ respectively, which may be due to the BEV features from the 2D multi-scale feature extractor containing more contextual information.

\begin{figure}[t]
  \begin{center}
    \includegraphics[width=\linewidth]{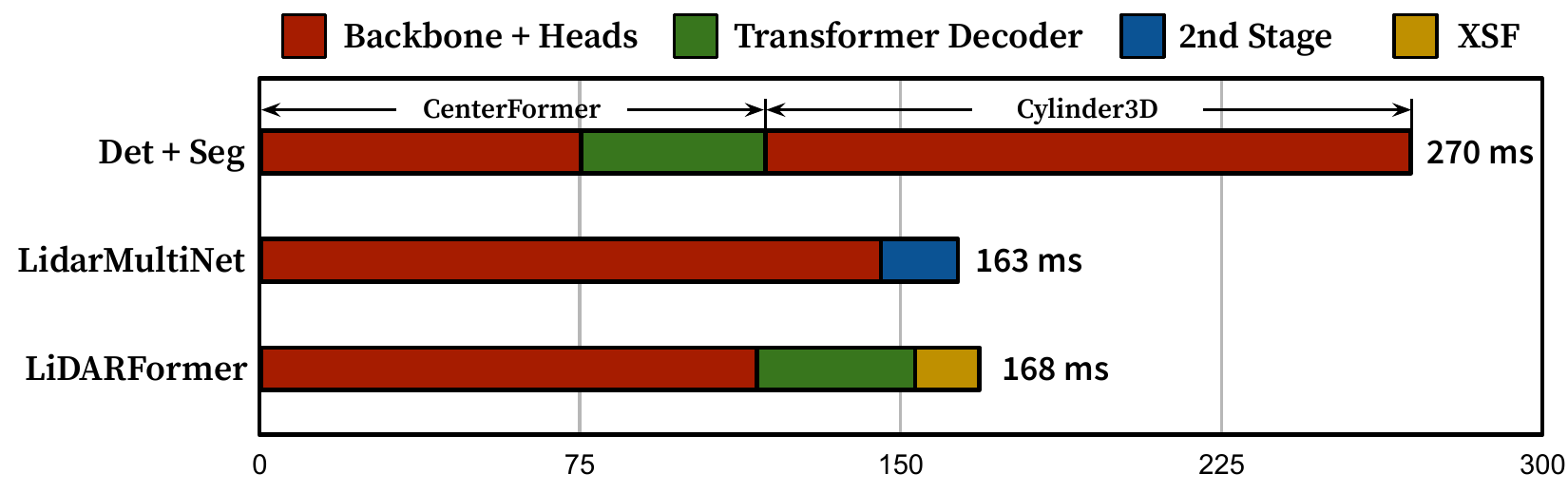}
  \end{center}
  \caption{\textbf{Inference Latency Comparison.} The notation ``Det + Seg'' refers to the combination of latencies from the previous SOTA detection and segmentation methods. Specifically, we chose CenterFormer~\cite{Zhou_centerformer} and Cylinder3D~\cite{zhu2021cylindrical} because they share similar backbone network structures with our approach. All methods were evaluated on Nvidia A100 GPU.}
  \label{fig:runtime}
\end{figure}

\begin{table}[t]
\centering
    \caption{The ablation of different query types adopted in the dense-to-sparse XSF on the nuScenes \texttt{val} split.}
    \label{table:query}
    \resizebox{0.8\linewidth}{!}{%
    \begin{tabular}{l|c}
    \Xhline{4\arrayrulewidth}
    LiDARFormer seg only result (mIoU) & 81.7\phantom{ (-1.7)} \\
    \Xhline{2\arrayrulewidth}
    Voxel Query & 81.4 (\textcolor{green}{-0.3}) \\
    Embedding Query  & 81.3 (\textcolor{green}{-0.4}) \\
    \Xhline{4\arrayrulewidth}
    \end{tabular}
    }
\end{table}
 
 \textbf{Runtime and Model Size}
We evaluated the runtime and model size of LiDARFormer on Nvidia A100 GPU. Figure~\ref{fig:runtime} demonstrates that a multi-task network can significantly reduce latency by sharing backbone networks. Our approach has similar latency to the previous 2-stage multi-task network but outperforms it in an end-to-end 1-stage network design. Additionally, LiDARFormer employs fewer parameters (77M) than the LidarMultiNet (131M).
 
 \begin{table}[t]
\centering
    \caption{Comparison of different class feature embedding initialization methods.}
    \label{table:init}
    \resizebox{0.8\linewidth}{!}{%
    
    \begin{tabular}{l|ccc}
    \Xhline{4\arrayrulewidth}
    Initialization Method & mAP & NDS & mIoU\\
    \Xhline{2\arrayrulewidth}
    BEV & \underline{66.6} & \underline{70.8} & 82.7\\
    Voxel & 66.4 & 70.5 & \underline{82.9}\\
    \Xhline{4\arrayrulewidth}
    \end{tabular}
    }
\end{table}

\begin{figure*}[t]
  \begin{center}
    \includegraphics[width=0.9\textwidth]{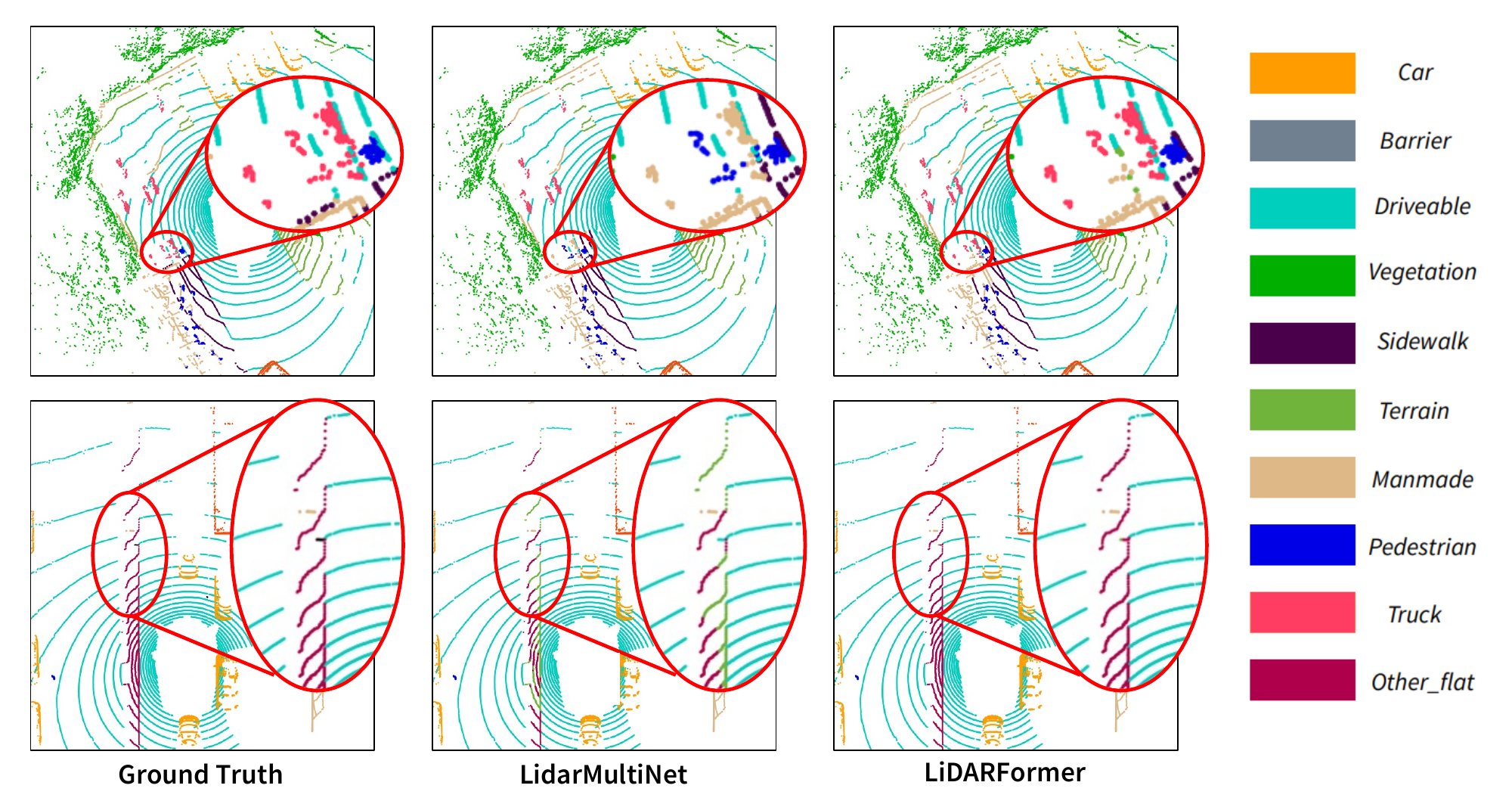}
  \end{center}
  \caption{\textbf{Visualization of the improvement of our model on the nuScenes \texttt{val} split.} With our proposed cross-space and cross-task transformer module, our approach generates more accurate labels in the previously uncertain area. Best viewed in color.}
  \label{fig:cmp_baseline_ours}
\end{figure*}

\begin{table*}[t]
	\centering
	\caption{Segmentation results of each class on the \texttt{test} split of nuScenes. We underline the best performance in each category.}
	\label{tab:nusc_seg_test_full}
	\resizebox{\linewidth}{!}{%
	\begin{tabular}{l|c|*{16}{c}}
		\Xhline{4\arrayrulewidth}
		Model & mIoU & \rotatebox{60}{barrier} &	\rotatebox{60}{bicycle} &	\rotatebox{60}{bus} &	\rotatebox{60}{car}&	\rotatebox{60}{\makecell{construction\\vehicle}} &	\rotatebox{60}{motorcycle} &	\rotatebox{60}{pedestrian} &	\rotatebox{60}{traffic cone} &	\rotatebox{60}{trailer} &	\rotatebox{60}{truck} & \rotatebox{60}{\makecell{driveable\\ surface}} &	\rotatebox{60}{other flat} &	\rotatebox{60}{sidewalk} &	\rotatebox{60}{terrain} &	\rotatebox{60}{manmade} &	\rotatebox{60}{vegetation} \\
		\Xhline{4\arrayrulewidth}
		PolarNet~\cite{Zhang_2020_CVPR} &  69.8 & 80.1 & 19.9 & 78.6 & 84.1 & 53.2 & 47.9 &  70.5 & 66.9 &  70.0 & 56.7 & 96.7 & 68.7 & 77.7 &  72.0 &  88.5 &  85.4 \\
		PolarStream~\cite{chen2021polarstream} &  73.4 & 71.4 & 27.8 & 78.1 & 82.0 & 61.3 &  77.8 & 75.1 &  72.4 & 79.6 & 63.7 & 96.0 & 66.5 &  76.9 &  73.0 &  88.5 & 84.8 \\
		JS3C-Net~\cite{yan2020sparse} &  73.6 & 80.1 & 26.2 & 87.8 & 84.5 & 55.2 &  72.6 & 71.3 &  66.3 & 76.8 & 71.2 & 96.8 & 64.5 &  76.9 &  74.1 &  87.5 & 86.1 \\
		Cylinder3D~\cite{zhu2021cylindrical} &  77.2 & 82.8 & 29.8 & 84.3 & 89.4 & 63.0 &  79.3 & 77.2 &  73.4 & 84.6 & 69.1 & 97.7 & 70.2 &  80.3 &  75.5 &  90.4 & 87.6 \\
		AMVNet~\cite{liong2020amvnet} &  77.3 & 80.6 & 32.0 & 81.7 & 88.9 & 67.1 &  84.3 & 76.1 &  73.5 & 84.9 & 67.3 & 97.5 & 67.4 &  79.4 &  75.5 &  91.5 & 88.7 \\
		SPVNAS~\cite{tang2020searching} &  77.4 & 80.0 & 30.0 & 91.9 & 90.8 & 64.7 &  79.0 & 75.6 &  70.9 & 81.0 & 74.6 & 97.4 & 69.2 &  80.0 &  76.1 &  89.3 & 87.1 \\
		Cylinder3D++~\cite{zhu2021cylindrical} &  77.9 & 82.8 & 33.9 & 84.3 & 89.4 & 69.6 &  79.4 & 77.3 &  73.4 & 84.6 & 69.4 & 97.7 & 70.2 &  80.3 &  75.5 &  90.4 & 87.6 \\
		AF2S3Net~\cite{Cheng_2021_CVPR} &  78.3 & 78.9 & \underline{52.2} & 89.9 & 84.2 & \underline{77.4} &  74.3 & 77.3 &  72.0 & 83.9 & 73.8 & 97.1 & 66.5 &  77.5 &  74.0 &  87.7 & 86.8 \\
		GASN~\cite{ye2022efficient} &  80.4 & 85.5 & 43.2 & 90.5 & 92.1 & 64.7 &  86.0 & 83.0 &  73.3 & 83.9 & 75.8 & 97.0 & 71.0 &  81.0 &  \underline{77.7} & 91.6 & \underline{90.2} \\
		SPVCNN++~\cite{tang2020searching} &  81.1 & \underline{86.4} & 43.1 & 91.9 & 92.2 & 75.9 &  75.7 & 83.4 &  77.3 & 86.8 & \underline{77.4} & 97.7 & \underline{71.2} &  81.1 &  77.2 & 91.7 & 89.0 \\
		LidarMultiNet~\cite{ye2022lidarmultinet} & 81.4 & 80.4 & 48.4 & \underline{94.3} & 90.0 & 71.5 &  87.2 & \underline{85.2} &  80.4 & 86.9 & 74.8 & 97.8 & 67.3 &  80.7 &  76.5 & 92.1 & 89.6 \\
		\Xhline{2\arrayrulewidth}
		LiDARFormer &  81.0 & 83.5 & 39.8 & 85.7 & 92.4 & 70.8 &  \underline{91.0} & 84.0 &  80.7 & \underline{88.6} & 73.7 & 97.8 & 69.0 & 80.9 & 76.9 & 91.9 & 89.0 \\
		LiDARFormer-TTA &  \underline{81.5} & 84.4 & 40.8 & 84.7 & \underline{92.6} & 72.7 &  \underline{91.0} & 84.9 &  \underline{81.7} & \underline{88.6} & 73.8 & \underline{97.9} & 69.3 & \underline{81.4} & 77.4 & \underline{92.4} & 89.6 \\
		
		\Xhline{4\arrayrulewidth}
	\end{tabular}
	}
\end{table*}
 
\textbf{Class Feature Embedding Initialization}
In our cross-task transformer decoder, we initialize the class feature embedding using a coarse prediction and its BEV features. As shown in Table~\ref{table:init}, if we change the initialization to use voxel features, the performance of LiDARFormer will increase in the segmentation task but will decrease in the detection task.

\textbf{Analysis of Cross-task Transformer}
We compare the segmentation prediction of our model to the baseline model without our proposed transformer module. As shown in Figure~\ref{fig:cmp_baseline_ours}, we notice that the improvement usually takes place in the discontinuous area, e.g. points from one object are predicted to have labels of different classes. 

\textbf{Polar Coordinate} PolarNet~\cite{Zhang_2020_CVPR} and Cylinder3D\cite{zhu2021cylindrical} have shown the potential of polar feature representation on the LiDAR segmentation problem. It mimics the scan pattern of the LiDAR sensor to balance the point distribution across different ranges of voxels. Contrary to their finding, the experiment on Waymo Open Dataset shows the performance on mIoU has a 1.4\% drop when we transform our baseline model to the polar coordinate with a similar voxel size. We conjecture that it is because we already use a relatively small voxel size, which does not induce a huge imbalance of points accumulation in the close-range voxel. Conversely, polar coordinates suffer more distortion in the distant voxels, leading to inferior performance.

\textbf{Class-wise Segmentation Results on nuScenes}
In Table~\ref{tab:nusc_seg_test_full}, we show the class-wise performance of LiDARFormer on the test set of nuScenes. The best segmentation results of each class are scattered among the top five methods. This is due to the learning competition among different classes. For example, better performance in ``motorcycle'' class will cause a drop in ``bicycle'' class. How to deal with the competition between similar classes is still an unsolved problem.

\begin{figure*}[th]
  \begin{center}
    \includegraphics[width=0.99\textwidth]{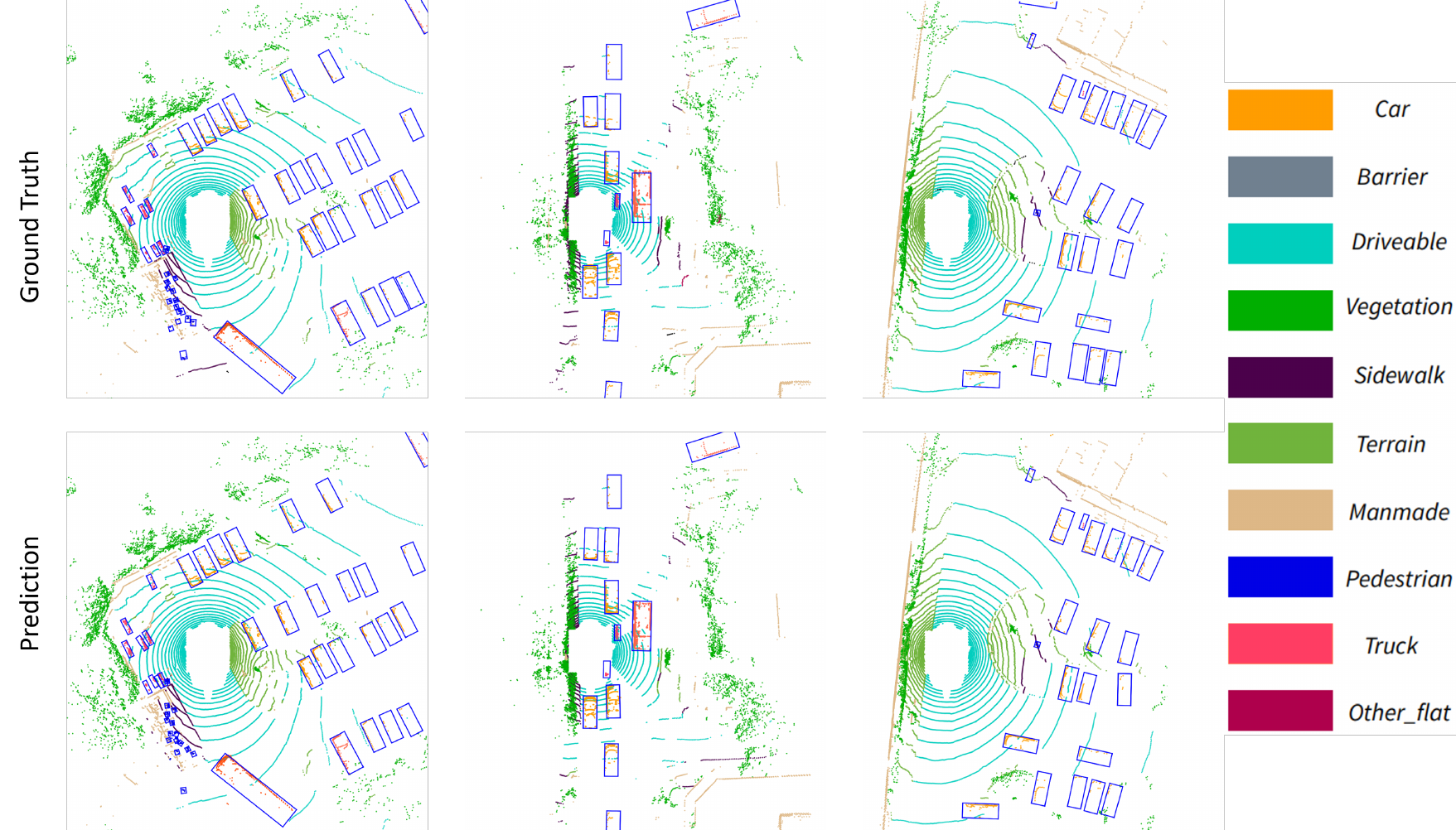}
  \end{center}
  \caption{Visualization of the detection and segmentation results on nuScenes.}
  \label{fig:nusc_vis}
\end{figure*}

\begin{figure*}[th]
  \begin{center}
    \includegraphics[width=0.99\textwidth]{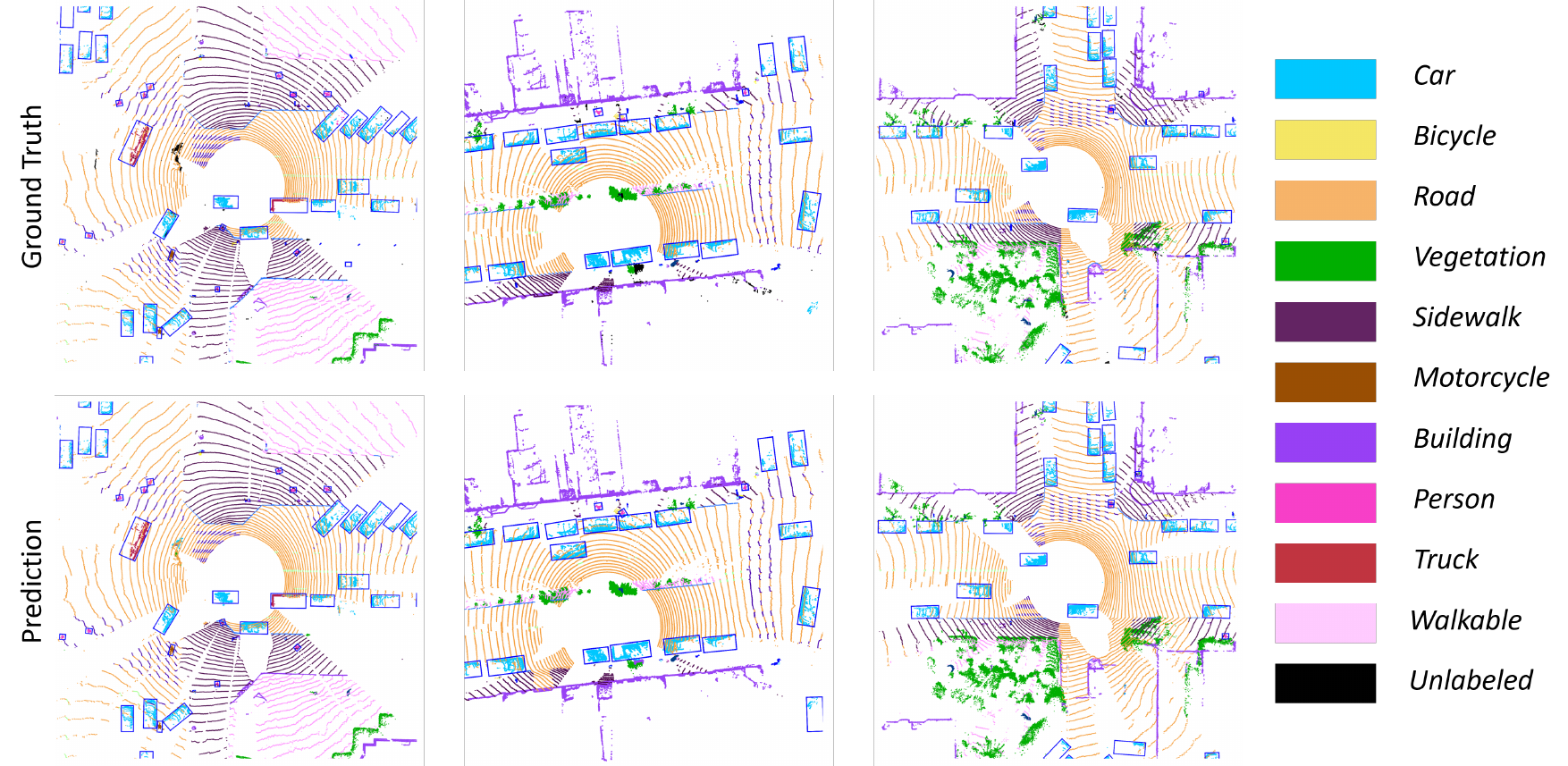}
  \end{center}
  \caption{Visualization of the detection and segmentation results on Waymo Open Dataset.}
  \label{fig:waymo_vis}
\end{figure*}

\section{Qualitative Results}
We illustrate the qualitative results of LiDARFormer on nuScenes and WOD in Figure~\ref{fig:nusc_vis},~\ref{fig:waymo_vis}. Our method can generate accurate semantic predictions in diverse environments.

{\small
\bibliographystyle{ieee_fullname}
\bibliography{citation}
}
\end{document}